\documentclass[11pt]{article}

\usepackage[preprint]{acl}

\usepackage{times}
\usepackage{latexsym}

\usepackage[T1]{fontenc}
\usepackage[utf8]{inputenc}

\usepackage{microtype}
\usepackage{algorithm}
\usepackage{algorithmic}

\usepackage{inconsolata}
\usepackage{amsmath}

\usepackage{graphicx}
\usepackage{algorithm}
\usepackage{algorithmic}

\usepackage{subcaption}
\usepackage{pifont}
\usepackage{xcolor}
\usepackage{tabularx}
\usepackage{booktabs}

\usepackage{booktabs}
\usepackage{multirow}
\usepackage{pifont}
\usepackage{xcolor}
\usepackage{array}

\title{Beyond Attention Magnitude: Leveraging Inter-layer Rank Consistency for Efficient Vision-Language-Action Models}

\author{
  Peiju Liu, Jinming Liu, Xipeng Qiu, \and Xuanjing Huang \\
  Fudan University \\
  \texttt{\{pjliu23, jmliu206\}@m.fudan.edu.cn} \\
  \texttt{\{xpqiu, xjhuang\}@fudan.edu.cn}
}

\begin{document}
\maketitle
\begin{abstract}
Vision-Language-Action (VLA) models excel in robotic manipulation but suffer from significant inference latency due to processing dense visual tokens. Existing token reduction methods predominantly rely on attention magnitude as a static selection. In this work, we challenge this assumption, revealing that high-attention tokens are task-dependent and can even degrade policy performance. To address this, we introduce \textbf{TIES} (\textbf{T}au-guided \textbf{I}nter-layer \textbf{E}fficient \textbf{S}election), a dynamic framework guided by inter-layer token ranking consistency. By adaptively balancing attention magnitude with ranking consistency, TIES ensures robust token selection without requiring additional training. On the CogACT + SIMPLER benchmark, TIES improves average success rates by 6\% while reducing token usage by 78\%, and demonstrates strong generalization across diverse decoders and benchmarks.
\end{abstract}

\section{Introduction}
Learning robust and generalizable policies for robotic manipulation has long been a fundamental challenge in policy learning. Traditional reinforcement learning methods often struggle with limited robustness and poor generalization in diverse scenarios. In recent years, the rapid progress of large-scale Vision-Language Models (VLMs) has showcased strong capabilities in multimodal reasoning and generalization. Building upon these advances, pioneering studies have proposed Vision-Language-Action (VLA) models, which fuse visual and linguistic inputs to directly produce robotic actions in an end-to-end fashion. While this paradigm significantly enhances the adaptability and generalization potential of robotic systems, it also introduces substantial computational overhead due to its quadratic nature in Transformer\cite{vaswani2017attention} when dealing with extended tokens.

To reduce the high computational cost of VLA models, previous work mainly relied on general-purpose acceleration techniques, including model lightweighting\cite{wen2025tinyvla}, quantization\citep{park2024quantization,fang2025sqap}, and early-exit mechanisms\cite{yue2024deer}. However, these strategies either largely overlook the intrinsic characteristics of VLA architectures, resulting in limited efficiency gains and marginal performance improvements, or require additional retraining procedures. More recently, some studies have identified visual tokens as a major computational bottleneck and proposed various token pruning approaches. Nevertheless, most of these methods are built upon the assumption that tokens with higher attention magnitude are inherently more important. Some approaches select high-attention tokens as anchors and progressively incorporate additional tokens based on heuristic rules such as geometric shapes, spatial overlap, or external detectors for objects and robot arms. These pipelines tend to be overly complex, computationally expensive, and lack interpretability.

\begin{figure}[t]
    \centering
    % 使用 \linewidth 确保子图宽度是相对于“当前栏”的宽度
    \begin{subfigure}[t]{0.49\linewidth}
        \centering
        \includegraphics[width=\linewidth]{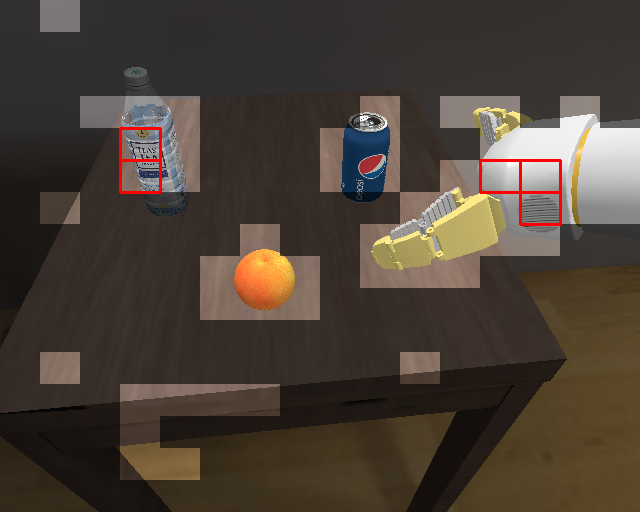}
        \subcaption{\textit{move bottle near orange}
        \textcolor{green}{\ding{52}} }
    \end{subfigure}
    \hfill % 自动填充中间压力，将两张图推向两边
    \begin{subfigure}[t]{0.49\linewidth}
        \centering
        \includegraphics[width=\linewidth]{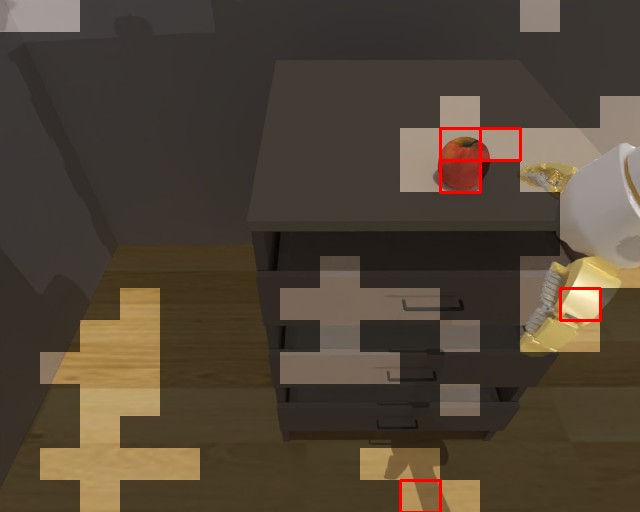}
        \subcaption{\textit{open middle drawer}
        \textcolor{red}{\ding{56}}}
    \end{subfigure}
    
    \caption{\textbf{Informative vs. misleading high-attention tokens.} Patches with high attention magnitude are unshaded, with the top 5 highlighted in red. (a) \textbf{Successful case:} high-attention tokens effectively localize task-relevant information. (b) \textbf{Failure case:} high-attention tokens focus on spurious features, leading to policy errors. Our approach introduces a consistency-based indicator to distinguish these scenarios and adaptively select tokens.}
    \label{fig:two_conditions}
\end{figure}

This study critically evaluates the widely adopted assumption that attention magnitude reliably indicates token importance. Our findings reveal that high-attention tokens are often task dependent and fluctuate significantly across states within the same task. We demonstrate that over-reliance on these tokens can degrade policy performance, indicating that attention magnitude lacks the sufficiency and robustness required for effective selection. Consequently, we introduce inter-layer token ranking consistency, a metric based on the Kendall rank correlation coefficient ($\tau$). Our results show that $\tau$ offers a more stable signal of token credibility across various tasks, architectures, and environments.

Building on this discovery, we propose TIES, a lightweight token pruning framework. TIES employs a small set of task images to establish a reference baseline distribution. During deployment, the framework dynamically adjusts the proportion of retained high-attention tokens by comparing the $\tau$ of the current frame to this distribution. To further reduce computational costs, TIES capitalizes on temporal redundancy in VLA inference. Pruning parameters are updated only upon detecting a significant visual shift between consecutive frames. This approach enables content aware, dynamic pruning that enhances inference speed without sacrificing control precision.

In summary, our contributions are as follows: \textbf{(1)~}We reveal that attention magnitude is an unreliable metric for token importance. \textbf{(2)}~We propose TIES, a novel framework that utilizes Kendall’s $\tau$ to measure inter-layer ranking consistency, enabling dynamic and content-aware token pruning. \textbf{(3)~}Our method reduces token usage by over 78\% while achieving a 6\% performance improvement over the full-token baseline using the CogACT model on the SIMPLER benchmark. Furthermore, we confirm its effectiveness across diverse VLA decoders and benchmarks.

\begin{figure*}[t] % 星号表示跨栏
    \centering
    \includegraphics[width=\textwidth]{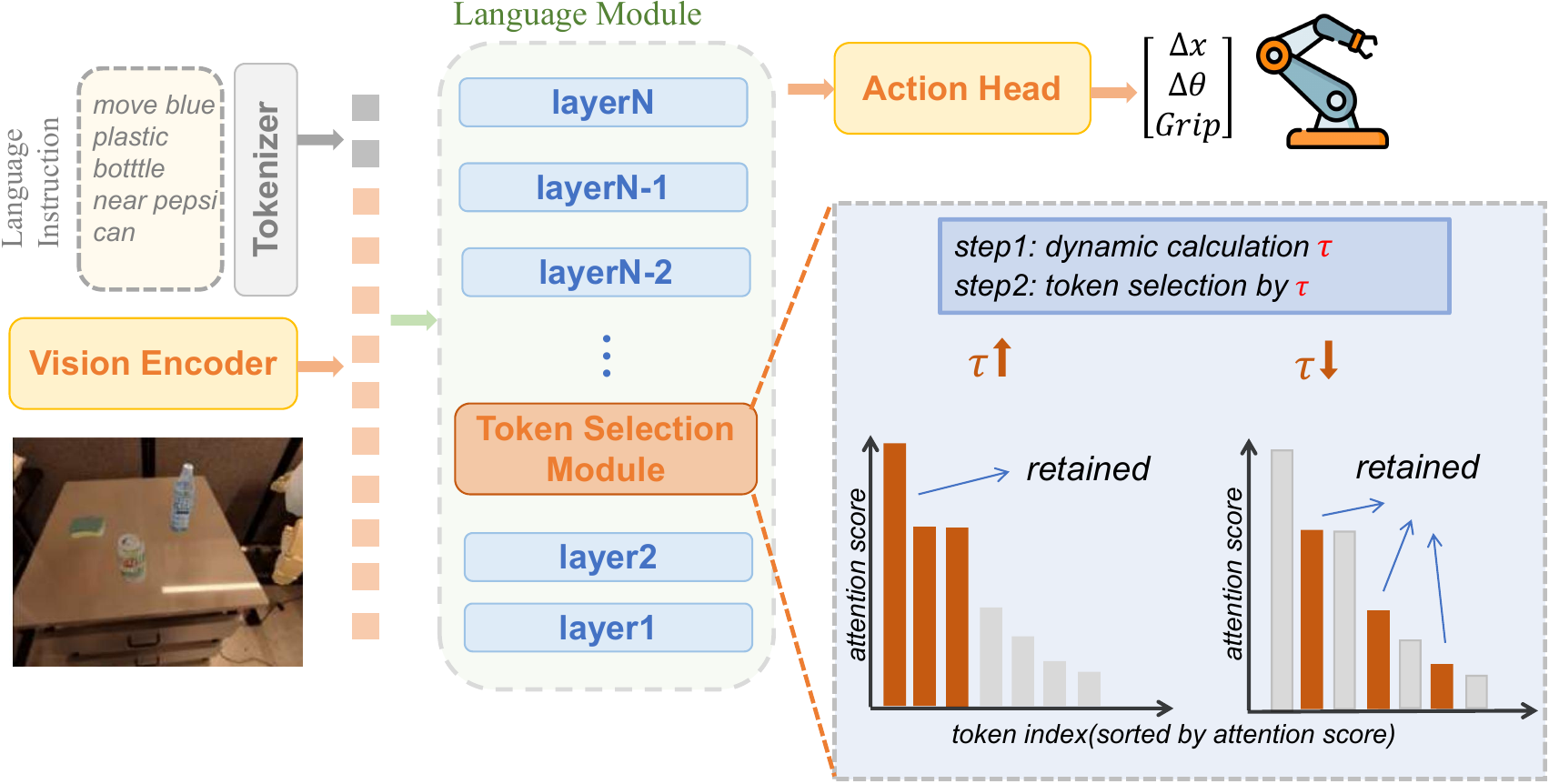}
    \caption{\textbf{TIES framework.} TIES dynamically computes the Kendall $\tau$ and adaptively decides the token selection strategy.}
\end{figure*}

\section{Related Work}
\subsection{Vision-Language-Action (VLA) Models}
VLA models typically leverage large language architectures fine-tuned on multimodal simulation~\cite{liu2023libero} and real-world datasets~\cite{o2024open}. Following the paradigm established by RT-2~\cite{zitkovich2023rt}  and OpenVLA~\cite{kim2024openvla}, recent models such as CogACT~\cite{li2024cogact} and $\pi_{0}$~\cite{black2410pi0} have demonstrated strong generalization in continuous action spaces. However, the quadratic complexity of self-attention in these large-scale models poses significant challenges for real-time deployment.

\subsection{VLA Acceleration and Token Pruning}
Previous research has explored various token-centric techniques to accelerate VLA models. For example, VLA-Cache ~\cite{xu2025vla} stores key-value pairs of visually similar or low-importance tokens from earlier steps and reuses them during subsequent inference. EfficientVLA ~\cite{yang2025efficientvla} attempts to identify task-relevant tokens using attention maps from a single transformer layer, combining techniques like layer pruning and decoder optimization. SP-VLA~\cite{li2025sp} proposes to preserve both spatial and semantic cues by retaining tokens with high feature saliency extracted from the vision encoder. SpecPrune-VLA~\cite{wang2025specprune} incorporates frame-level similarity and enhanced pruning strategies to improve inference efficiency. However, these methods are largely driven by heuristic design choices, lacking systematic analysis and comprehensive validation across diverse tasks and environments. 

Besides token reduction, other orthogonal techniques have been proposed to accelerate VLA inference~\cite{yu2025survey}. Spec-VLA~\cite{wang2025spec} introduces a speculative framework designed to accelerate VLA models. FlashVLA~\cite{tan2025think} proposes a
token-aware action reuse strategy by carefully checking prior actions with metrics of action and token stability. SQAP-VLA~\cite{fang2025sqap} proposes
new quantization-aware token pruning criteria that work on an aggressively quantized model to enhance pruning effectiveness. These methods focus on the decoding process and are compatible with our visual token selection framework, potentially offering synergistic efficiency gains.

\section{Key Insights}
\subsection{Priliminary}
Vision-Language-Action (VLA) models extend large-scale Vision-Language Models (VLMs) to enable end-to-end robotic control from multimodal inputs. These models are typically trained on large-scale datasets such as Open X-Embodiment~\cite{o2024open}, which contains over one million real-world robot trajectories. In a standard pipeline, a pre-trained vision encoder extracts visual features, which are then projected into the embedding space of a Large Language Model (LLM). The LLM processes the concatenated multimodal tokens and outputs action sequences. To enhance inference efficiency and policy expressivity, various decoder designs have been investigated, ranging from auto-regressive models (e.g., OpenVLA \citealp{kim2024openvla}) and linear regression heads (e.g., OpenVLA-OFT \citealp{kim2025fine}) to diffusion-based decoders (e.g., CogACT \citealp{li2024cogact}). In this paper, we demonstrate the versatility of our approach by validating it across these representative decoder architectures.

The primary computational bottleneck in VLA inference arises from the self-attention mechanism. Let the language instruction be tokenized into $N_l$ tokens and the visual observation into $N_v$ tokens. The total sequence length $N = N_l + N_v$ results in a computational and memory complexity of $O(N^2)$. Since $N_v$ is typically much larger than $N_l$ to preserve spatial resolution, the quadratic cost is dominated by visual tokens. To achieve real-time control, effective token reduction strategies are essential. Most existing pruning methods adopt attention magnitude as the primary criterion for importance. Given the attention matrix $A$, where $A_{i,j}$ denotes the attention weight from token $i$ to token $j$, an importance score $S_j$ is typically assigned by aggregating the attention values a token receives: $S_j = \sum_{i=1}^{N} A_{i,j}$. Tokens are then ranked by $S_j$, and only the top-$k$ tokens are retained.

While early works like FastV~\cite{chen2024image} established the paradigm of prioritizing high-attention tokens, recent VLM research has introduced additional dimensions for selection, specifically importance and diversity~\cite{alvar2025divprune,wen2025stop}. Importance is typically quantified by attention scores, whereas diversity is measured by feature distance or redundancy metrics. Although deduplication and distance-based sampling have been proven effective in VLMs~\cite{zhang2025beyond}, VLA research remains heavily reliant on attention-based ranking. Building on this, some recent efforts have incorporated specialized priors, such as object geometry or human-defined saliency\cite{li2025sp}. However, these approaches often rely on task-specific heuristics and maintain the underlying assumption that tokens with high attention weights are inherently more valuable for task execution. As we demonstrate in the next section, this unconditional reliance on attention magnitude is not always optimal in the complex context of robotic manipulation.

\subsection{Negative Effects of High-Attention Tokens}
Our investigation is motivated by recent discussion regarding the interpretability of attention mechanisms. While conventional wisdom suggests that tokens with high attention weights are semantically critical, recent studies offer a more nuanced perspective. For instance, ~\citet{xiao2023efficient} suggests that high-attention tokens primarily serve as sinks for information flow during the Transformer's forward pass to maintain numerical stability, rather than being purely task-relevant. Similarly, ~\citet{zhang2024cls} observed that attention maps in cross-modal Transformers are often noisier and less reliable proxies for token importance compared to those in standalone vision encoders. Furthermore, ~\citet{darcet2023vision} argues that vision encoders disperse semantic information across multiple patches and often assign high weights to "register" tokens that lack semantic meaning. This renders individual visual tokens uninterpretable to human cognition and casts doubt on the validity of simple saliency heuristics.

\begin{figure}[t]
  \centering % 确保图片在缩窄后依然居中
  \includegraphics[width=0.94\columnwidth]{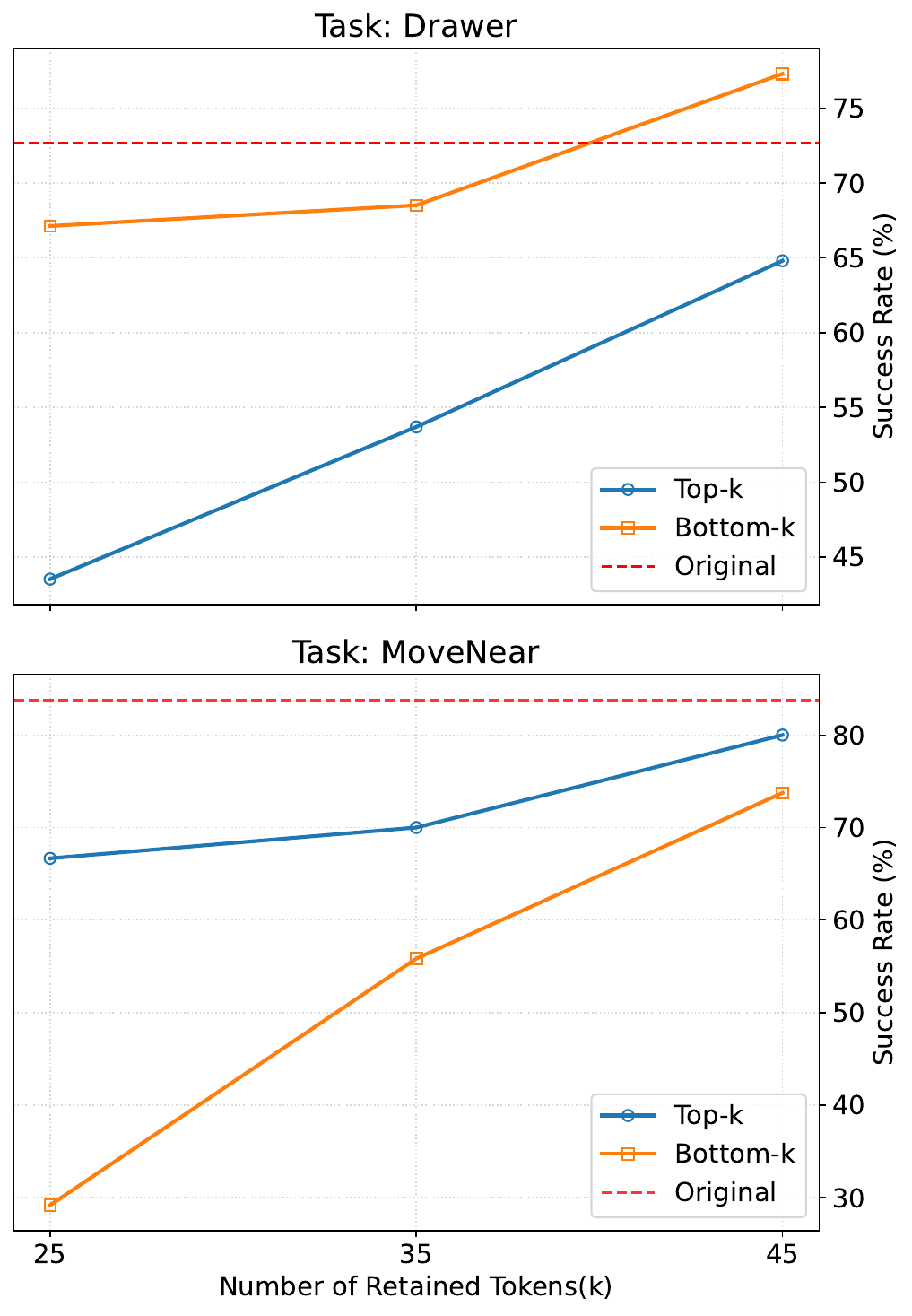}
  \caption{\textbf{Performance of Top‑k and Bottom‑k strategies}. We observe performance inversion in the \textit{Drawer} task while conventional attention importance holds in the \textit{MoveNear} task}
  \label{fig:pos_neg}
\end{figure}

We challenge the presumed synergy between high attention and token importance through a counter-intuitive experiment. Specifically, we evaluate the performance impact of retaining a reduced set of tokens under two extreme regimes: \textbf{Top-$k$}, where we only keep $k$ tokens with the highest attention scores, and \textbf{Bottom-$k$}, where we retain a subset of tokens excluding the highest-ranked ones (specifically, we keep the last $k$ tokens among the top $K$). We compare these against the original baseline with full 256 tokens reserved. As shown in Figure \ref{fig:pos_neg}, the role of high-attention tokens varies drastically across different robotic tasks. 

\textbf{Performance Inversion.} In the \textit{Drawer} task, we observe a striking logic inversion. The Bottom-45 strategy achieves a success rate of 77.31\%, which not only significantly outperforms the Top-45 strategy but also exceeds the full-token Baseline by a margin of 4.61\%. This suggests that in certain manipulation tasks, the highest-ranked tokens can act as spurious signals or noise that misleads the model's policy and to remove them effectively serves as a denoising mechanism. 

\textbf{Task Dependency.} Conversely, in the \textit{MoveNear} task, the conventional assumption holds true. Top-45 achieves 80.0\%, clearly outperforms Bottom-45 with 73.75\%. Furthermore, as $k$ decreases from 45 to 25, the performance of the Bottom-k strategy in \textit{MoveNear} collapses more rapidly compared to the \textit{Top-k} strategy, indicating a higher reliance on primary attention signals for this specific task.

These results demonstrate that the credit or reliability of high-attention tokens is highly task-dependent and can even vary across different states within the same task. Blindly relying solely on magnitude as a selection criterion is insufficient and can be detrimental to policy performance.

\subsection{Identifying Indicators for Attention Reliability}
To develop an adaptive token selection mechanism, the primary challenge lies in identifying a reliable metric that distinguishes whether high-attention tokens are beneficial (Type 1) or detrimental (Type 2) to policy performance.

\begin{figure}[t]
\centering
  \includegraphics[width=0.98\columnwidth]{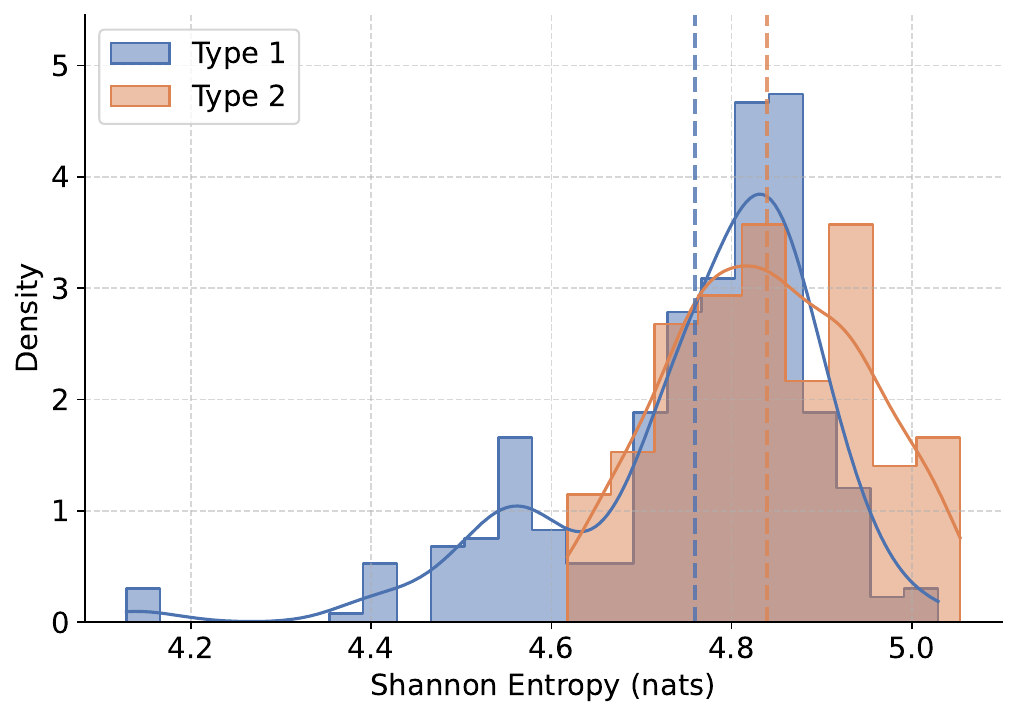}
  \caption{\textbf{Entropy distribution.}}
  \label{fig:entropy}
\end{figure}

\textbf{Limitations of Static Statistical Metrics.} A natural starting point is to examine the attention distribution's sharpness using Shannon Entropy. One might hypothesize that low-entropy distributions represent more confident and thus more reliable token selection. However, our analysis on the SIMPLER benchmark refutes this. As shown in Figure \ref{fig:entropy}, the entropy distributions for both Type 1 and Type 2 scenarios exhibit massive overlap. The mean entropy for both types converges closely, making it statistically impossible to separate reliable frames from unreliable ones using magnitude-based or distribution-based static metrics alone.

\textbf{From Static Magnitude to Dynamic consistency.} Inspired by the observation that token importance is not static but flows and evolves across Transformer layers\cite{zhang2024cls}, we shift our focus from absolute weights to the inter-layer rank dynamics. We track the rank fluctuations of the top-ranked tokens during the forward pass. Preliminary qualitative analysis suggests that the indices of high-attention tokens are highly volatile, and the nature of this volatility differs between the two different scenarios. To quantify this, we employ the Kendall rank correlation coefficient ($\tau$) to measure the rank consistency of the top-$k$ tokens between consecutive layers. Given two rankings, the consistency between them is defined as:$$\tau_0 = \frac{P - Q}{\sqrt{(P+Q+T)(P+Q+U)}}$$where $P$ is the number of concordant pairs, $Q$ is the number of discordant pairs. We define $T$ and $U$ as the number of tied pairs within the first and second ranking sequences, respectively. Our final consistency metric $\tau$ is reported as the layer-wise average of $\tau_0$ calculated between successive layers. A higher $\tau$ indicates that the model's focus remains a relatively more stable set of tokens across layers.

\begin{figure}[h]
\centering
  \includegraphics[width=0.98\columnwidth]{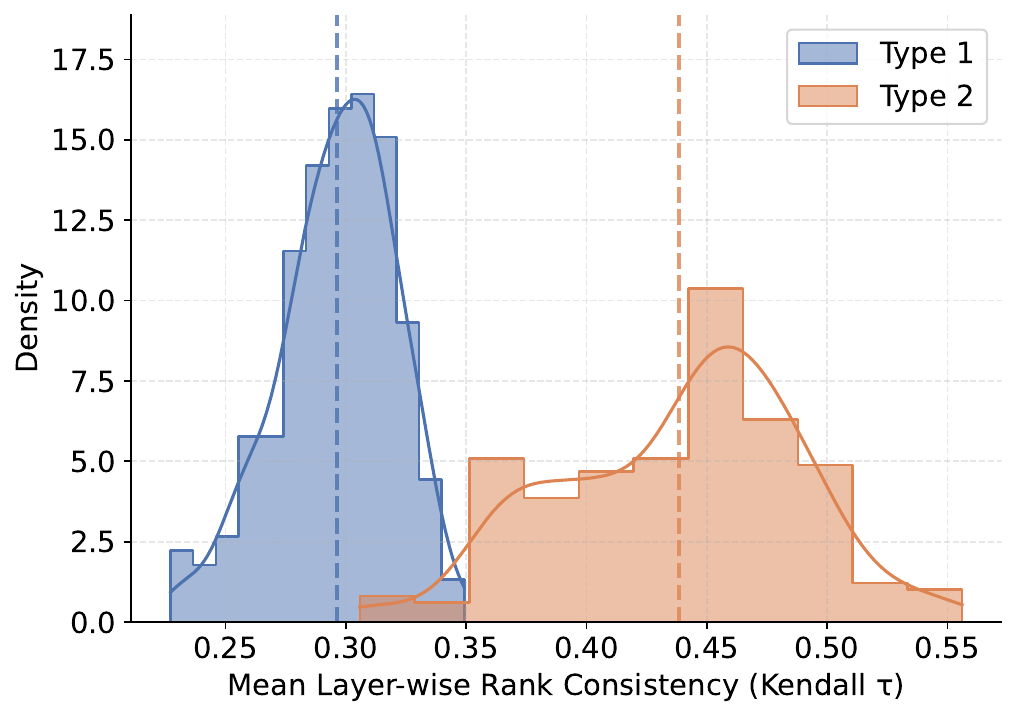}
  \caption{\textbf{Kendall $\tau$ distribution.}}
  \label{fig:tau_credit}
\end{figure}

\textbf{The Divergence of Rank Consistency.} Our quantitative analysis reveals a striking divergence in rank consistency between the two types. As illustrated in Figure \ref{fig:tau_credit}, the Kendall $\tau$ distributions for Type 1 and Type 2 are effectively decoupled: Type 1 exhibits significantly lower consistency, indicating a more dynamic reassessment of token importance as the representation evolves.Type 2 exhibits high consistency, where the attention rankings remain rigidly fixed across the depth of the model. Statistical evaluation shows a strong negative correlation between $\tau$ and task types, achieving an AUC of 0.91 in predicting attention reliability.

\textbf{The Spurious Locking Hypothesis.} Our finding presents a compelling counter-intuitive insight: consistency is a sign of failure, while volatility is a sign of health. We hypothesize that in the scenarios where we should not trust attention, the model suffers from spurious correlation locking. When $\tau$ is high, the attention mechanism is likely trapped in a local optimum, fixating on redundant or harmful features (e.g., background textures) that were prioritized in early layers and fail to be refined. Conversely, a lower $\tau$ suggests a healthy information flow where the model iteratively updates its focus based on the evolving context of the task. By utilizing $\tau$ as a lightweight, training-free indicator, we can dynamically switch between aggressive attention-based pruning and diversity-preserving sampling, forming the basis of our TIES framework.

\section{Method}
The core of our approach is TIES, a training-free dynamic token pruning framework. As illustrated in Algorithm \ref{alg:prune}, TIES adaptively modulates the reliance on high-attention tokens based on the inter-layer rank consistency ($\tau$) of the input.

\subsection{$\tau$-guided Token Selection}
We utilize the Kendall rank correlation coefficient $\tau$ as a dynamic indicator to distinguish whether high-attention tokens are beneficial or detrimental to policy performance. We propose two selection strategies within the module to balance Importance (attention-driven) and Informativeness (representational coverage). Hard-TIES is a threshold-based binary selection. If $\tau > \tau_{threshold}$, the model identifies the current state as Type 2, where attention is likely trapped in spurious correlations. In this case, it switches entirely to alternative strategy like Uniform sampling to ensure diversity. Otherwise, it defaults to Top-$k$ Sampling. Soft-TIES offers a more granular control using linear interpolation. We define a weight $w_t = \text{Interpolate}(\tau_t, \tau_{med})$, which determines the proportion of tokens retained from different sources. Specifically, the model first selects a subset of tokens from the top-ranked candidates based on attention scores, then supplements the remaining required tokens using Uniform Sampling or other diversity-based algorithm like DivPrune\cite{alvar2025divprune} to maintain representational coverage.

\begin{algorithm}[t]
\caption{$\tau$-guided Dynamic Token Pruning (TIES)}\label{alg:prune}
\begin{algorithmic}[1]
\REQUIRE Input sequence $I=\{i_1, i_2, \dots\}$, Prune ratio $\rho$, VLA Model $M$, Similarity threshold $\gamma$, Total tokens $N$
\STATE \textbf{Phase 1:} Calibration (Offline)
\STATE $\mathcal{D} \leftarrow$ Sample $M$ frames from diverse trajectories
\STATE Calculate $\{\tau_i\}_{i=1}^{M}$ to determine median $\tau_{med}$ and distribution $\Sigma$
\STATE \textbf{Phase 2:} Dynamic Inference (Online)
\STATE $\tau_{curr} \leftarrow \text{None}$, $i_{anchor} \leftarrow \text{None}$
\FOR{each timestep $t = 1, 2, \dots$}
    \IF{$i_{anchor}$ is None \OR $\text{Sim}(i_t, i_{anchor}) < \gamma$}
        \STATE $i_{anchor} \leftarrow i_t$ 
        \STATE $\tau_{curr} \leftarrow$ compute Kendall $\tau$ for $i_t$ across layers of $M$
        \STATE $w_t \leftarrow \text{Interpolate}(\tau_{curr}, \tau_{med})$ \COMMENT{Calculate trust weight}
    \ENDIF
    \STATE $N_{top} = \lfloor w_t \cdot \rho N \rfloor$, $\quad N_{uni} = \rho N - N_{top}$
    \STATE $\mathcal{T}_{top} \leftarrow$ Select top $N_{top}$ tokens based on $\sum A_{i,j}$
    \STATE $\mathcal{T}_{uni} \leftarrow$ Uniformly sample $N_{uni}$ tokens from remaining set
    \STATE $\mathcal{T}_{final} = \mathcal{T}_{top} \cup \mathcal{T}_{uni}$
    \STATE Execute forward pass of $M$ using $\mathcal{T}_{final}$
\ENDFOR
\end{algorithmic}
\end{algorithm}

\subsection{Implementation Details}
To minimize the computational overhead of calculating $\tau$ in high-frequency control loops, we leverage the temporal redundancy of robotic observations. Instead of a fixed-interval update, we implement a Similarity-based Trigger: The first frame of a temporal sequence is designated as an anchor, and its $\tau$ is computed to set the strategy. For subsequent frames, the model monitors the visual similarity against the anchor. A new $\tau$ is only recomputed when the similarity drops below a threshold $\gamma$. This ensures the pruning strategy remains robust to scene changes while avoiding redundant calculations for static observations and minimal changes. In addition, to ensure the robustness of the $\tau$ indicator, we perform a one-time offline calibration. We randomly sample $M$ frames from the target dataset to estimate the distribution of ranking variations $\Sigma$ and the median $\tau_{med}$. This allows TIES to be transferable to new environments within the same domain. In our practice, $M=100$ yields sufficiently decent performance.

\section{Experiment}
\subsection{Experimental Settings}
\textbf{Simulation Implementation Details.}~To evaluate our approach, we mainly conduct experiments on the SIMPLER\cite{li2024evaluating} environment, a simulation-based benchmark specifically designed for table-top manipulation tasks. SIMPLER aims to faithfully reproduce real-world robotic dynamics for platforms such as the Google Robot and WidowX, and has been shown to exhibit strong sim-to-real consistency. In this setup, the VLA model receives 224×224 RGB visual observations along with natural language task descriptions (e.g., "Pick coke can") as inputs, and generates a sequence of actions in a 7-DoF Cartesian control space. SIMPLER offers two evaluation protocols: Visual Matching, which emphasizes visual consistency with real-world scenes, and Variant Aggregations, which introduces a variety of environmental variations including changes in lighting, backgrounds, and surface textures. For the Google robot, both protocols are supported, and each includes the same set of four manipulation tasks: (1) Pick coke can; (2) Move near; (3) Open/close drawer; and (4) Open the top drawer and place the apple. We adopt task success rate as the primary evaluation metric.

\textbf{Baselines.}~We conduct our primary experimental validation of TIES on the CogACT\cite{li2024cogact} framework, which features a sophisticated architecture incorporating DINOv2\cite{oquab2023dinov2} and SigLIP\cite{zhai2023sigmoid}) as vision encoders, a Llama2-7B\cite{touvron2023llama} backbone for multimodal reasoning, and a Diffusion Transformer~\cite{peebles2023scalable} for action trajectory synthesis. To evaluate its performance, we compare TIES against several state-of-the-art baselines, including: FastV\cite{chen2024image}, which enhances inference speed via redundant visual token pruning; VLA-Cache\cite{xu2025vla}, which employs temporal analysis to skip redundant computations by caching static tokens; and EfficientVLA\cite{yang2025efficientvla}, a comprehensive approach integrating token pruning, layer reduction, and decoder optimization.

\textbf{Implementation Details.}~In all experiments, the pruning process was consistently applied from the second Transformer layer. All our emperical evaluations were executed on NVIDIA A800 GPUs.

\subsection{Main Results on SIMPLER}
\begin{table*}
\centering
\resizebox{\textwidth}{!}{%
\renewcommand{\arraystretch}{1.2}
\begin{tabular}{l l c c c c c}
\toprule
\textbf{SIMPLER} & \textbf{Method} &
\textbf{PickCan} & \textbf{MoveNear} & \textbf{Drawer} &
\textbf{DrawerApple} & \textbf{Average} \\
\midrule
\multirow{6}{*}{\textbf{Visual Matching}}
& CogACT & 91.7  & 83.8  & 72.7  & 42.6  & 72.7   \\
& FastV  & 92.6    & 81.4    & 69.8   & 52.4   & 74.1    \\
& VLA-Cache & 92.0  & 83.3   & 70.5   & 51.6   & 74.4    \\
& EfficientVLA & 94.7  & 82.4  & 69.8 &55.4 & 75.5  \\
\cmidrule(lr){2-7}
& Soft-TIES & \underline{93.3} & 79.6 & \underline{76.9} & \underline{57.4} & \underline{76.8} \\
& Hard-TIES & 91.3 & 79.2 & \textbf{77.1} & \textbf{64.8} & \textbf{78.1} \\
\midrule
\multirow{6}{*}{\textbf{Variant Aggregation}}
& CogACT & 88.6  & 76.5  & 28.8  & 45.6  & 59.9   \\
& FastV  & 91.4    & 78.6    & 27.6   & 50.6   & 62.1    \\
& VLA-Cache & 91.7  &79.3   & 32.5   & 45.8   & 62.3    \\
& EfficientVLA & 94.4  & 77.2  & 27.6 &51.3 & 62.6  \\
\cmidrule(lr){2-7}
& Soft-TIES & \underline{88.7} & 76.2 & \textbf{44.3} & \textbf{61.1} & \textbf{67.6} \\
& Hard-TIES & \underline{89.7} & 74.7 & \underline{41.0} & \underline{54.6} & \underline{65.0} \\
\bottomrule
\end{tabular}
}
\caption{Performance of TIES compared with other baselines on the SIMPLER environment. All pruning methods are evaluated with 56 retained tokens.}\label{tab:simpler}
\end{table*} 

We evaluate the performance of TIES using the SIMPLER benchmark across two distinct protocols: Visual Matching (VM) and Variant Aggregation (VA). As summarized in Table \ref{tab:simpler}, TIES consistently outperforms existing baselines and the full-token CogACT model in all evaluated scenarios. 

In the Visual Matching protocol, which emphasizes consistency with real-world scenes, Hard-TIES achieves a superior average success rate of 78.1\% while retaining only 56 tokens. This performance represents a 5.4\% improvement over the unpruned CogACT baseline and significantly exceeds the results of competitive acceleration methods such as FastV and VLA-Cache. The robustness of our framework is further highlighted in the Variant Aggregation protocol, where environmental perturbations such as lighting and texture changes typically degrade policy reliability. In this challenging setting, the unpruned CogACT baseline experiences a sharp decline in performance to 59.9\%. In contrast, Soft-TIES maintains a much higher average success rate of 67.6\%, effectively bridging the gap between efficiency and robustness. 

Task-specific analysis reveals even more striking insights into the limitations of standard attention-driven selection. In the Visual Aggregation \textit{Drawer} task, the baseline achieves a 28.8\% success rate, while Soft-TIES improves this to 44.3\%. This huge performance inversion, that fewer tokens lead to better results, directly supports our hypothesis that high-attention tokens can sometimes act as spurious sinks or noise. By utilizing the Kendall $\tau$ coefficient to detect these instances, TIES selectively trust the tokens with high magnitude, ensuring that the robotic policy remains focused on task-relevant features rather than redundant visual artifacts.

\section{Ablation Study}
\begin{figure}
  \includegraphics[width=\columnwidth]{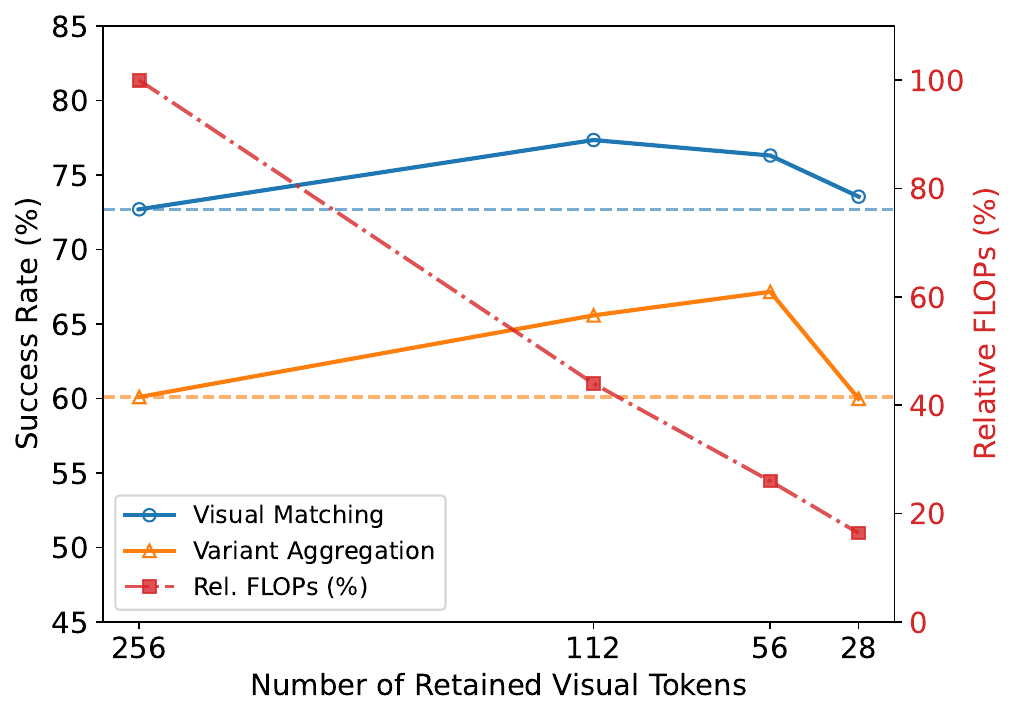}
  \caption{\textbf{Impact of pruning ratio.}}
  \label{fig:pruning_ratio}
\end{figure}
\subsection{Impact of Pruning Ratio}
                                   
Figure \ref{fig:pruning_ratio} illustrates the trade-off between policy performance and computational efficiency as the number of retained tokens $T$ varies across $\{28, 56, 112, 256\}$. Our TIES framework demonstrates remarkable robustness to aggressive pruning , achieving the peak performance at $T=112$ where the success rate in Visual Matching increases from 72.7\% to 77.35\%, and in Variant Aggregation reaches 67.16\% at $T=56$. This performance inversion suggests that $\tau$-guided selection effectively prunes noisy or spurious visual signals that would otherwise mislead the policy. Even at $T=28$, representing an 83.6\% reduction in FLOPs, the success rate still surpasses the unpruned baseline. By dynamically pivoting to alternative strategy in untrustful states, TIES acts as a denoising filter that enhances generalization under environmental perturbations.

\begin{figure}[h]
  \includegraphics[width=\columnwidth]{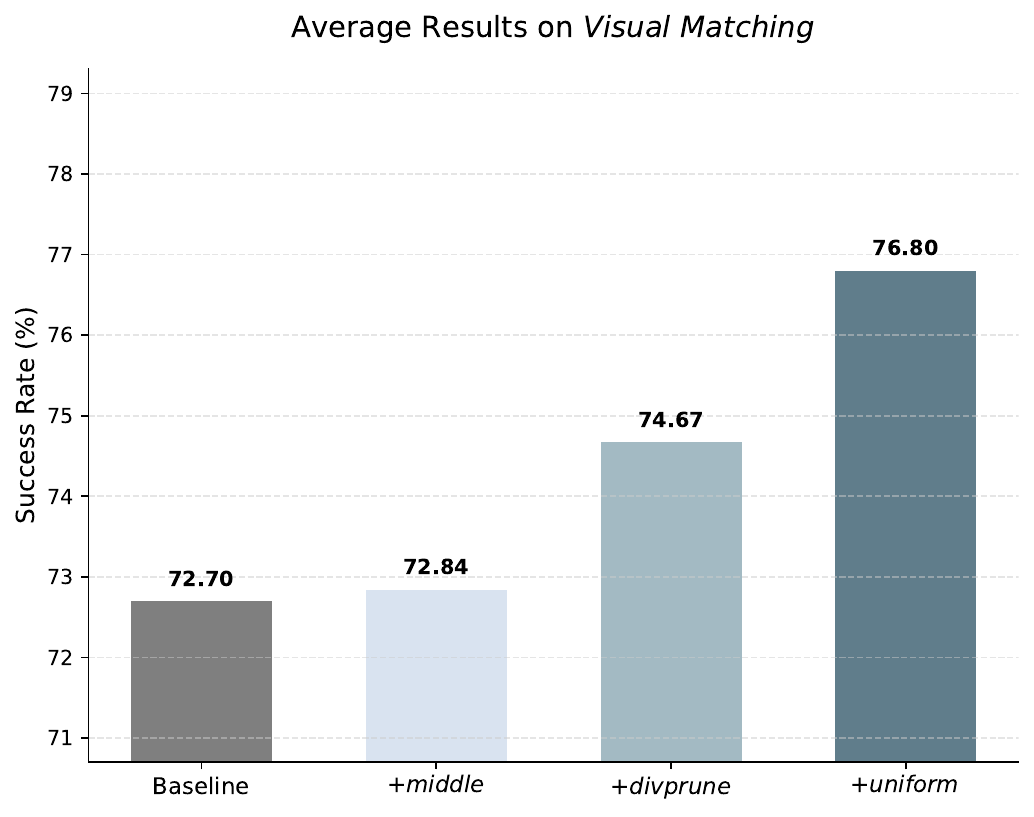}
  \caption{\textbf{Comparisons of alternative strategies.} All strategies yield consistent performance gains.}
  \label{fig:strategy}
\end{figure}

\subsection{Comparisons of Alternative Strategies}

Figure \ref{fig:strategy} shows the average performance of different token selection strategies on the Visual Matching task with 56 tokens retained. Specifically, \textbf{Middle} strategy keeps tokens with median attention scores, while \textbf{Uniform} samples tokens uniformly according to their attention ranking. \textbf{Divprune}~\cite{alvar2025divprune} is used as a strong baseline that considers token redundancy and deduplication. Notably, all these strategies outperform the full-token baseline even under a high pruning ratio. Interestingly, the Uniform strategy achieves the best performance despite its simplicity. These results demonstrate that with the $\tau$ indicator to distinguish different scenarios, even simple heuristic methods can effectively maintain stable and strong performance.

\subsection{Generalization Across Architectures and Benchmarks}
To assess the robustness of our method across different architectures and environments, we evaluate its performance using OpenVLA~\cite{kim2024openvla} as the base model on the SIMPLER benchmark. Unlike the CogACT framework, OpenVLA employs an auto-regressive decoder that requires multiple forward passes per action, making inference efficiency particularly critical. As shown in Table~\ref{tab:average_results}, our method consistently outperforms the FastV counterpart, with notable gains in visual aggregation scenarios. Furthermore, on the widely adopted LIBERO benchmark~\cite{liu2023libero}, we evaluate the fine-tuned OpenVLA-OFT~\cite{kim2025fine} model. Our approach again yields superior results compared to FastV, demonstrating its effectiveness and generalizability when transferred to different decoders and task domains.

\begin{table}[htbp]
\centering
\small % 稍微缩小字号
\begin{tabular}{@{}lccc@{}}
\toprule
~ & \textbf{SIMPLER VM} & \textbf{SIMPLER VA} & \textbf{LIBERO} \\ \midrule
Full tokens & 34.30 & 39.30 & 96.90 \\
\midrule
FastV & 27.99 & 31.47 & 92.70 \\
\textbf{TIES} & \underline{28.31} & \textbf{39.44} & \underline{94.30} \\ \bottomrule
\end{tabular}
\caption{\textbf{Average success rate comparison across benchmarks.} All pruning methods are evaluated with 56 retained tokens.}
\label{tab:average_results}
\end{table}

\section{Conclusion}
In this work, we show that relying on tokens with high magnitude is insufficient and occasionally detrimental to efficient VLA-based robotic manipulation. Instead, We utilize inter-layer token rank consistency as a more reliable indicator. Building on this insight, we develop TIES, a dynamic token-selection strategy that balances attention magnitude and alternative cues. Experiments on diverse model architectures and benchmarks demonstrate that TIES successfully achieves high performance while maintaining computational efficiency in robotic manipulation.

\section*{Limitations}
While TIES effectively mitigates attention traps and enhances inference efficiency, its logic relies on a heuristic correlation between inter-layer rank consistency and token importance, lacking deep theoretical grounding in the causal chain of model decision-making. Furthermore, the current selection strategy primarily focuses on filtering spatial redundancy, leaving room to explore more sophisticated cross-modal alignment that could further refine the interaction between visual features and complex linguistic instructions.

\bibliography{custom}

\end{document}